\documentclass[letterpaper, 10 pt, conference]{ieeeconf}  

\IEEEoverridecommandlockouts                              

\usepackage{graphicx} 
\usepackage{graphicx} 
\usepackage{amsmath} 
\usepackage{amssymb}  
\usepackage{amsfonts} 
\usepackage{booktabs}
\usepackage{graphicx}
\usepackage[table]{xcolor}
\usepackage[skins,breakable]{tcolorbox}
\tcbuselibrary{listings}

\usepackage{listings}
\usepackage{tcolorbox}
\usepackage{pifont}
\usepackage{multirow}
\usepackage{float}
\usepackage{placeins}
\usepackage{caption}
\let\labelindent\relax
\usepackage{enumitem}
\newcommand{\cmark}{\ding{51}}
\newcommand{\xmark}{\ding{55}}
\newcommand{\method}{AtomVLA}

\title{\LARGE \bf
AtomVLA: Scalable Post-Training for Robotic Manipulation via Predictive Latent World Models
}

\author{
Xiaoquan Sun$^{1,3}$, Zetian Xu$^{1,2}$, Chen Cao$^{1,2}$, Zonghe Liu$^{1,2}$, Yihan Sun$^{3}$,
Jingrui Pang$^{4}$,\\ Ruijian Zhang$^{3}$, Zhen Yang$^{1,2}$, Kang Pang$^{3}$, Dingxin He$^{3}$,
Mingqi Yuan$^{1,2}$, Jiayu Chen$^{1,2}$\\[1ex]
{\small $^{1}$INFIFORCE Intelligent Technology Co., Ltd. Hangzhou, China}\\
{\small $^{2}$The University of Hong Kong, Hong Kong SAR, China}\\
{\small $^{3}$Huazhong University of Science and Technology, Wuhan, China}\\
{\small $^{4}$Tsinghua University, Beijing, China}\\[0.5ex]
{\small Corresponding author: Jiayu Chen, \texttt{jiayuc@hku.hk}}
}

\begin{document}
\twocolumn[{%
\renewcommand\twocolumn[1][]{#1}%
\maketitle
\maketitle
\thispagestyle{empty}
\pagestyle{empty}
\begin{center}
    \centering
    \includegraphics[width=0.99\linewidth]{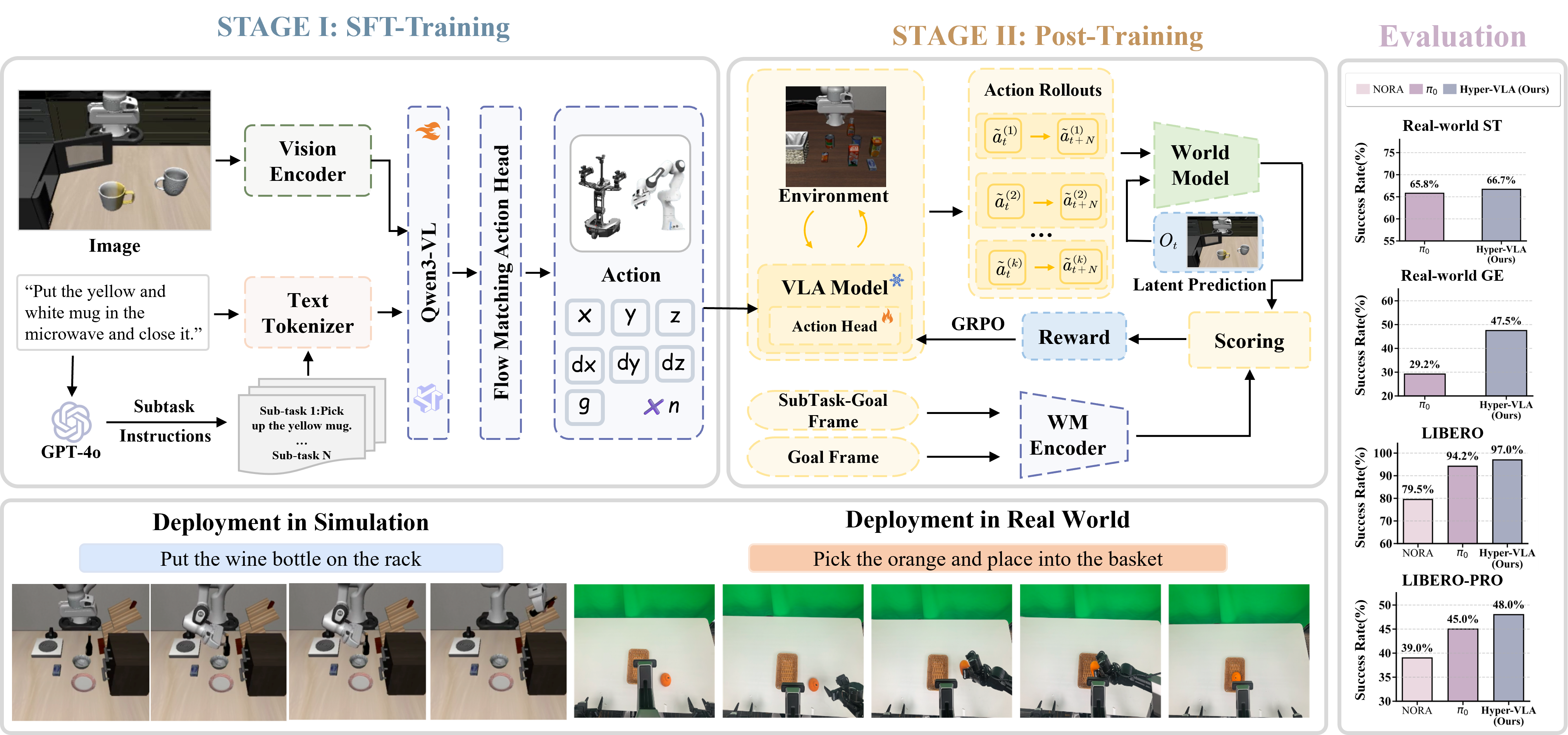}
    \captionsetup{font=footnotesize}
    \captionof{figure}{\textbf{Framework of \method.} We propose a scalable two-stage framework for robotic manipulation. \textbf{Left (Stage I):} high-level instructions are decomposed into subtask instructions using a Large Language Model (GPT-4o). Subsequently, these subtask instructions are integrated alongside the original high-level instruction as guidance for the SFT training of the model. \textbf{Middle (Stage II):} A predictive latent world model evaluates candidate action rollouts to provide reward for offline post-training via GRPO. \textbf{Right \& Bottom:} \method~ achieves 97\% and 48\% success rates on LIBERO and LIBERO-PRO benchmarks and demonstrates strong generalization in real-world.}
    \label{fig:teaser}
    \vspace{1.5em}
\end{center}
}]

\begin{abstract}
Vision-Language-Action (VLA) models demonstrate remarkable potential for generalizable robotic manipulation. The execution of complex multi-step behaviors in VLA models can be improved by robust instruction grounding, a critical component for effective control. However, current paradigms predominantly rely on coarse, high-level task instructions during supervised fine-tuning. This instruction grounding gap leaves models without explicit intermediate guidance, leading to severe compounding errors in long-horizon tasks. Therefore, bridging this instruction gap and providing scalable post-training for VLA models is urgent. To tackle this problem, we propose \method, the first subtask-aware VLA framework integrated with a scalable offline post-training pipeline. Our framework leverages a large language model to decompose high-level demonstrations into fine-grained atomic subtasks.  This approach utilizes a pretrained predictive world model to score candidate action chunks against subtask goals in the latent space, mitigating error accumulation while significantly improving long-horizon robustness.
Furthermore, this approach enables highly efficient Group Relative Policy Optimization without the prohibitive expenses associated with online rollouts on physical robots. Extensive simulations validate that our AtomVLA maintains strong robustness under perturbations. When evaluated against fundamental baseline models, it achieves an average success rate of 97.0\% on the LIBERO benchmark and 48.0\% on the LIBERO-PRO benchmark. Finally, experiments conducted in the real world using the Galaxea R1 Lite platform confirm its broad applicability across diverse tasks, especially long-horizon tasks. All datasets, checkpoints, and code will be released to the public domain following the acceptance of this work for future research.
\end{abstract}


\section{INTRODUCTION}
Vision-Language-Action (VLA) models have significantly advanced the development of embodied intelligence, offering a unified framework for end-to-end visuomotor control and complex instruction-following~\cite{reconvla,intelligence2025pi_,kim2024openvla,udvla}. By leveraging the reasoning capabilities of large-scale multi-modal models, they directly interpret natural language
instructions into executable robot actions, demonstrating remarkable potential for generalizable robot manipulation. However, the transition from lab-scale demonstrations to robust long-horizon deployment is consistently constrained by several critical challenges. Specifically, current VLA architectures~\cite{nora15,pi0} demand prohibitive computational resources and massive trajectory datasets for pre-training, making them difficult to scale in resource-constrained environments. Furthermore, the semantic-visual information asymmetry is widespread across these models, as high-dimensional visual tokens often dominate the latent space. As a result, the learned policy struggles to ground sparse linguistic instructions into sequential physical guidance, leading to compounding errors in multi-step tasks. Finally, the high costs and safety risks of real-world interaction make online reinforcement learning (RL) impractical for most robotic platforms, leaving a gap between static imitation learning and dynamic policy refinement.

To address these challenges, effort has been devoted to computational overhead optimization \cite{llava-vla,smolvla,tinyvla}, semantic reasoning improvement \cite{cyclevla}, and the reduction of physical interaction costs through generative simulation \cite{RoboTwin,tabletopgen}. Specifically, LLaVA-VLA~\cite{llava-vla}, SmolVLA~\cite{smolvla} and TinyVLA~\cite{tinyvla}~focuses on building lightweight models to significantly reduce parameter counts; CycleVLA~\cite{cyclevla} introduces cyclic mechanisms to enhance semantic comprehension; and RoboTwin~\cite{RoboTwin,tabletopgen} generates extensive interaction data via simulation environments. Despite these significant advancements, existing methods often suffer from poor generalization in practical deployment due to their reliance on imitation learning. Furthermore, most current frameworks focus on reactive action generation, lacking the latent understanding and prediction of real-world dynamic evolution. This absence of modeling capability hinders the model from performing scalable offline policy optimization in the absence of costly online feedback.

In this paper, we propose \textbf{AtomVLA}, a novel, scalable, two-stage post-training framework for robust VLA models in long-horizon robotic tasks. Our contributions are threefold:
\begin{itemize}
    \item AtomVLA first introduces a robust architecture that bridges the instruction-grounding gap by leveraging a large language model (LLM) to decompose complex, high-level demonstrations into fine-grained atomic subtasks. This decomposition provides explicit stage-wise guidance that aligns the policy's semantic understanding with its underlying action-chunking mechanism. 
    \item Furthermore, to eliminate reliance on costly real-world interactions and to bypass the hallucinations inherent in generative models, a predictive latent world model based on V-JEPA2~\cite{vjepa2} is employed to rigorously evaluate candidate action trajectories against subtask goals directly in the latent space. Equipped with these latent transitions, the framework generates highly reliable reward signals for RL post-training, enabling efficient policy refinement without the computational overhead or visual artifacts of pixel-level synthesis.
    \item We demonstrate state-of-the-art performance, achieving a 97.0\% success rate on the LIBERO benchmark alongside robust visual generalization with a 48.0\% success rate on the LIBERO-PRO benchmark. Furthermore, we verify the physical reliability of the model on the Galaxea R1 Lite platform, specifically highlighting its success in complex long-horizon manipulation of deformable objects such as folding a T-shirt.
\end{itemize}

\section{RELATED WORK}
\subsection{Vision-Language-Action Models} 
To achieve better adaptability and efficacy, early works~\cite{rt1,pi0} optimized Transformer architectures from the ground up, leveraging massive internet-sourced multimodal datasets alongside extensive robotic execution traces. Building upon this foundation, OpenVLA~\cite{kim2024openvla} released the first open source foundation model trained on large-scale public data. Subsequently, OpenHelix~\cite{openhelix} proposed a dual system architecture for robotic manipulation. To improve deployment efficiency in practical environments, PD-VLA~\cite{PD-VLA} and CEED-VLA~\cite{CEED-VLA} explored methods to accelerate the inference process. Furthermore, recent studies such as Spatial Forcing~\cite{spatialforcing} and SpatialVLA~\cite{spatialvla} have investigated the incorporation of three dimensional spatial features. However, these approaches oriented toward three dimensional perception rely excessively on visual representations at the pure image level. Consequently, they exhibit significant limitations regarding the deep grounding of natural language instructions and the design of their overall training paradigms.

\subsection{Reinforcement Learning for Post-Training}
Reinforcement learning post-training has emerged as a crucial paradigm to overcome the inherent limitations of supervised fine-tuning~\cite{WoVR,wmpo,worldvla,simplevla-rl,motus}. Because directly deploying online reinforcement learning on physical robots incurs prohibitive interaction costs~\cite{sop}, recent research increasingly favors adopting world models as virtual simulators. Numerous studies~\cite{world-env,wmpo,vla-rft,matrix-game2} have explored this intersection by utilizing simulated environments to generate synthetic experiences for policy optimization. However, effectively fusing these two components remains a significant challenge. Standard generative world models typically synthesize future states at the pixel level. This mechanism inevitably leads to the accumulation of autoregressive errors and produces visual hallucinations during long sequence predictions~\cite{WoVR}.
To address these bottlenecks, we introduce a scalable post-training pipeline that combines a predictive latent world model (V-JEPA) with GRPO to score trajectories and refine policies. This yields a reliable reward signal and stable improvements on long-horizon tasks.
\begin{figure}[t]
    \centering
    \includegraphics[width=1.0\linewidth]{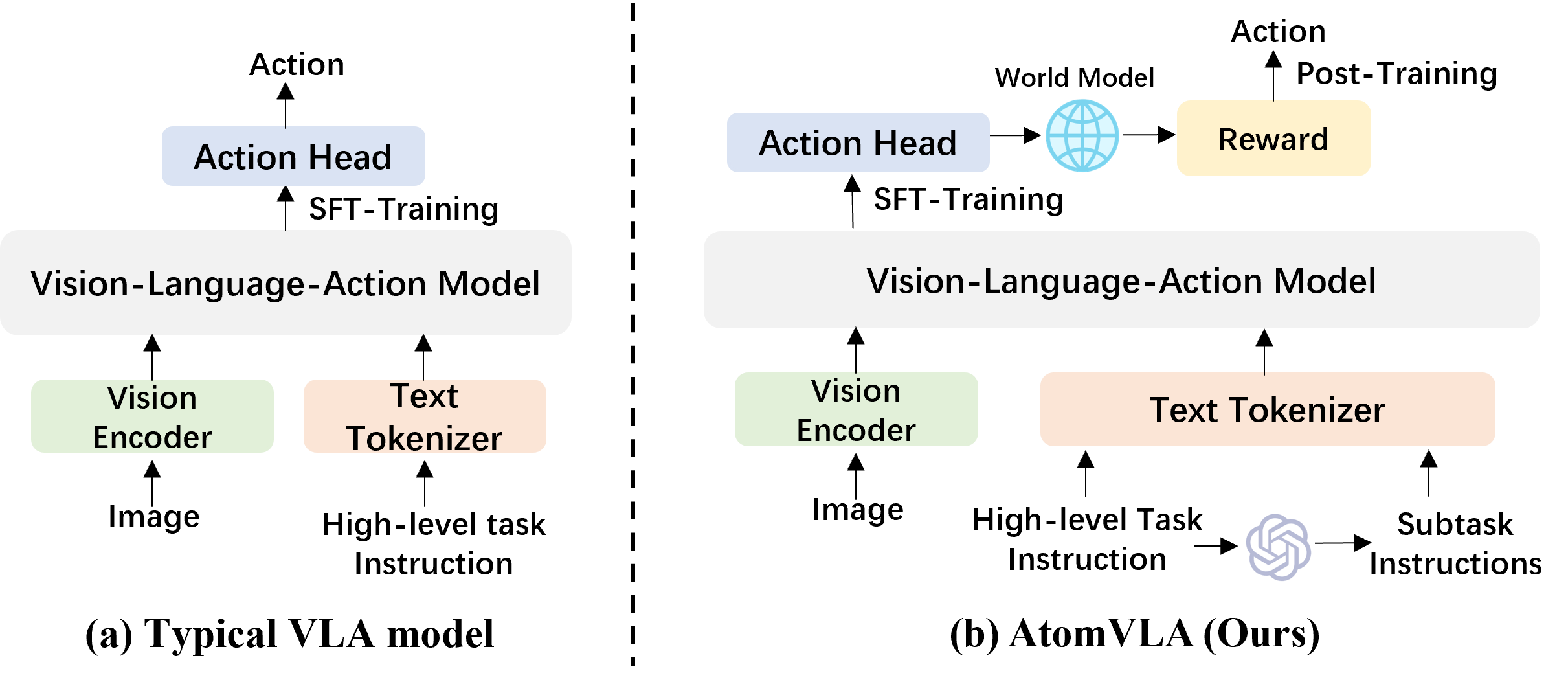}
\caption{(a) Typical VLA models rely on SFT Training. (b) AtomVLA (Ours) leverages a language model for fine-grained decomposition of atomic subtask instructions and a world model for RL post-training.}
    \label{fig:platform}
\end{figure}

\begin{figure*}[!t]
    \centering
    \includegraphics[width=0.9\linewidth]{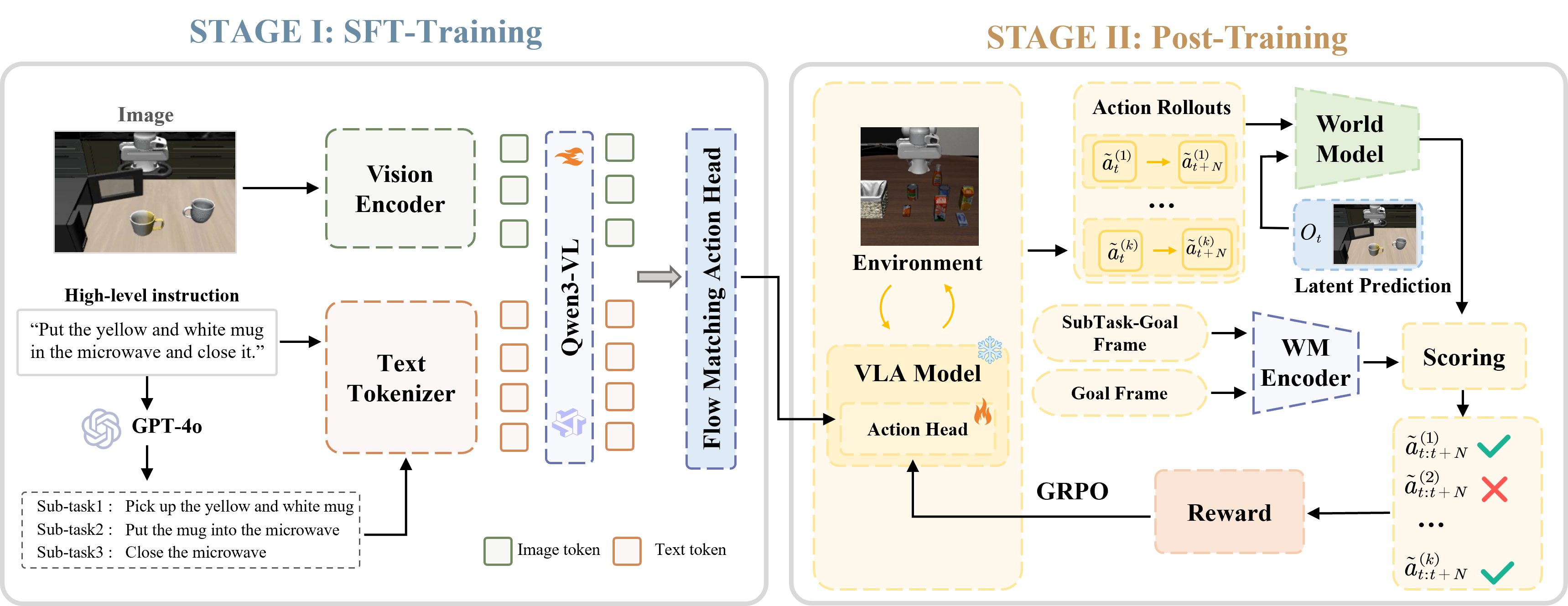}
    \caption{\textbf{Training pipeline.} \textbf{Stage I:} high-level instructions are decomposed into fine-grained atomic subtask instructions using LLM (GPT-4o). Subsequently, these subtask instructions are integrated with the original high-level instruction to guide the SFT training of the model. \textbf{Stage II:} A predictive latent world model evaluates candidate action rollouts to provide reward for offline post-training. }
    \label{fig:training_pipline}
\end{figure*}

\section{\method}
\subsection{Model Architecture}

\noindent\textbf{VLM Backbone.}
\method~uses the open-source multi-modal Qwen3-VL-4B-Instruct~\cite{Qwen3-VL} as the VLM backbone and couples it with an action head.
At time $t$, given a multi-view visual observation $O_t$, high-level instruction $I_t$, and subtask instruction $SI_t$, the backbone encodes them into contextual token features
\begin{equation}
\mathbf{H}_t = f_{\mathrm{VLM}}(O_t, I_t+ SI_t).
\end{equation}
which condition the action expert via cross-attention.
The expert outputs an action chunk $\mathbf{a}_{t:t+N}=[\mathbf{a}_t,\ldots,\mathbf{a}_{t+N}]$, where $N$ is the action horizon.

\noindent\textbf{Action Head.}
We introduce an action head, implemented as a cross-attention Diffusion Transformer~\cite{DiT}.
Specifically, the action expert serves as a flow-matching head that regresses an action sequence.
Conditioned on $\mathbf{H}_t$ and $S_t$, the expert predicts a horizon-$N$ action chunk
$\mathbf{a}_{t:t+N}$. Following a flow-matching formulation, we corrupt the ground-truth chunk with Gaussian noise:
sample $\boldsymbol{\epsilon}\sim\mathcal{N}(0,\mathbf{I})$ and a continuous time $\tau\in(0,1)$, then form
$\mathbf{x}_\tau=(1-\tau)\boldsymbol{\epsilon}+\tau\mathbf{a}_{t:t+N}$.
The expert is trained to regress the target velocity $\mathbf{v}=\mathbf{a}_{t:t+N}-\boldsymbol{\epsilon}$ by predicting
$\hat{\mathbf{v}}=\mathcal{A}_\phi(\mathbf{Gx}_\tau,\tau;\mathbf{H}_t,s_t)$ and minimizing
\begin{equation}
\mathcal{L}_{\mathrm{FM}}=\mathbb{E}_{\tau,\boldsymbol{\epsilon}}
\left\|\mathcal{A}_\phi(\mathbf{x}_\tau,\tau;\mathbf{H}_t,s_t)-(\mathbf{a}_{t:t+N}-\boldsymbol{\epsilon})\right\|_2^2.
\label{eq:flow_matching_loss}
\end{equation}
At inference time, we initialize $\mathbf{a}^{(0)}\sim\mathcal{N}(0,\mathbf{I})$ and iteratively refine it for $K$ steps
using explicit Euler updates $\mathbf{a}^{(k+1)}=\mathbf{a}^{(k)}+\Delta t\,\mathcal{A}_\phi(\mathbf{a}^{(k)},\tau_k;\mathbf{H}_t,s_t)$,
with $\Delta t = 1/K$, yielding the final action chunk.

\noindent\textbf{Subtask Dataset.}
We use a large language model (LLM) to further decompose each demonstration trajectory and its high-level task instruction into fine-grained atomic subtasks, and to annotate each subtask with a natural language instruction and the start and end frames. 
Given the high-level task instruction, we provide the LLM with a sampled sequence of video frames and the task context, and ask it to segment the entire demonstration into a number of atomic subtasks. 
The LLM returns a JSON list of tuples in the format $\{(\ell_i, s_i, e_i)\}_{i=1}^{M}$, where $i$ indexes subtasks, $M$ is the number of subtasks, $\ell_i$ is the subtask instruction, and $s_i, e_i$ are the start and end frames. To ensure consistent granularity, we explicitly constrain the prompt so that the generated subtasks align with a small set of basic manipulation actions, rather than overly fine-grained intermediate process descriptions. For example, for the \emph{pick and place} task, we standardize subtasks into action expressions such as \texttt{Pick up [object]}, \texttt{Place [object] on [target position]}, \texttt{Open/Close [object]}, and \texttt{Push [object]}. Under this rule, a task is decomposed into 2-5 subtasks. Prompt details are provided in the supplementary material. In our implementation, we use GPT-4o as the LLM for fine-grained atomic subtask decomposition and annotation.

\subsection{Reward Modeling and RL Post-training}
\label{sec:reward_grpo}
\noindent\textbf{Motivation.}
While SFT on demonstrations equips the policy with a strong imitation capability, it often falls short on more challenging tasks and generalizes poorly.
A common remedy is to further apply reinforcement learning post-training to improve long-horizon decision making and error recovery; however, collecting online RL rollouts in simulators or on real robots is costly and difficult to scale.
We therefore seek a scalable offline post-training signal that encourages goal-directed progress at an appropriate temporal granularity and remains stable by anchoring learning to expert demonstrations.
To this end, we treat subtask boundary frames as intermediate goals and leverage a pretrained world model to score candidate action chunks, enabling scalable RL post-training.

\noindent\textbf{Action-conditioned world model.}
We build an action-conditioned latent dynamics model on top of a pretrained video representation.
Concretely, we adopt V-JEPA2~\cite{vjepa2} as a frozen visual encoder $J(\cdot)$ that maps an image to a latent token sequence.
Given the current observation $O_t$ and a candidate action chunk $\tilde{\mathbf{a}}_{t:t+N}$, a predictor network $W_\theta$ rolls out the latent future
$\hat{z}_{t+N}=W_\theta(J(O_t),\tilde{\mathbf{a}}_{t:t+N})$.
This enables scoring candidate actions by comparing their predicted consequences against goal states in latent space.

\noindent\textbf{Reward design.}
We decompose each episode into coarse-grained subtasks and obtain boundary frames for each subtask. For a timestep $t$, let $b(t)$ denote the boundary frame index of the current subtask, i.e., the first frame after the current subtask ends.
We treat $O_{b(t)}$ as the \emph{subgoal} and $O_{T-1}$ as the final task goal within the same episode.
For each candidate $\tilde{\mathbf{a}}^{(k)}_{t:t+N}$, we compute two goal energies
\[
E_{\text{sub}}^{(k)}=\left\|W_\theta\!\left(J(O_t),\tilde{\mathbf{a}}^{(k)}_{t:t+N}\right)-J(O_{b(t)})\right\|_1,
\]
\vspace{2pt}
\[
E_{\text{goal}}^{(k)}=\left\|W_\theta\!\left(J(O_t),\tilde{\mathbf{a}}^{(k)}_{t:t+N}\right)-J(O_{T-1})\right\|_1.
\]
where lower is better.
To discourage implausible actions and prevent reward hacking, we also include an imitation deviation term
$D^{(k)}=\|\tilde{\mathbf{a}}^{(k)}_{t:t+N}-\mathbf{a}^\star_{t:t+N}\|_1$.
The final scalar reward is
\[
r^{(k)}=-\Big(\lambda_{\text{sub}}E_{\text{sub}}^{(k)}+\lambda_{\text{goal}}E_{\text{goal}}^{(k)}+\alpha D^{(k)}\Big).
\]

Intuitively, the subgoal term provides stage-wise guidance aligned with subtask progress, while the final-goal term enforces long-horizon consistency, where $\lambda_{\text{sub}} = 0.3$, $\lambda_{\text{goal}} = 0.4$, and $\alpha = 0.3$.

\noindent\textbf{GRPO optimization.}
Given $K$ candidates per state, we normalize rewards within each group to form advantages $A^{(k)}$ and update the policy to increase the likelihood of higher-advantage candidates.
We further regularize updates with a KL penalty to a frozen SFT reference policy $\pi_{\mathrm{ref}}$ for stability:
\begin{equation}
\label{eq:grpo_objective}
\begin{aligned}
\mathcal{L}_{\mathrm{grpo}}
&= -\mathbb{E}_{\mathbf{H}_t}\!\Bigg[ \frac{1}{K} \sum_{k=1}^{K} \left( A^{(k)}
\log \pi_\phi\!\left(\tilde{\mathbf{a}}^{(k)}_{t:t+N}\mid \mathbf{H}_t \right) \right) \\
&\qquad\qquad - \lambda\,\mathrm{KL}\!\left(\pi_\phi\,\|\,\pi_{\mathrm{ref}}\right) \Bigg].
\end{aligned}
\end{equation}
We only update the action head parameters during post-training.

\subsection{Training Pipeline}

As shown in Fig.~\ref{fig:training_pipline}, there are two major training stages:
\noindent\textbf{SFT Training.}
We perform supervised fine-tuning on robot demonstrations by jointly optimizing the Qwen3-VL backbone and the action expert.
Given $(O_t, I_t, S_t)$, we supervise the predicted action chunk $\mathbf{a}_{t:t+N}$ using the flow-matching loss in Eq.~\eqref{eq:flow_matching_loss}.
We further reduce gradient variance by repeating each batch with independently sampled $(\tau,\boldsymbol{\epsilon})$ via the repeated diffusion steps and training with mixed precision.



\noindent\textbf{Post Training.}
Starting from the SFT checkpoint, we conduct offline GRPO post-training using the reward model defined in Sec.~\ref{sec:reward_grpo}.
Concretely, we first sample states from the same offline demonstration dataset used in Stage I. 
Initialized from the Stage I checkpoint, the current policy then generates multiple candidate action chunks for each state. 
Next, we compute subgoal and final goal rewards with the V-JEPA2~\cite{vjepa2} world model, and perform GRPO updates with a KL constraint to the frozen SFT reference policy.
Unless otherwise specified, we mainly update the action expert parameters during post-training to preserve stable vision-language representations while improving long-horizon robustness.

\section{SIMULATION EXPERIMENTS}
We concentrate on several experiments to answer the following questions:
\begin{itemize}[leftmargin=*]
\item\textbf{Q1}: Does \method{} outperform existing VLA models on standard and long-horizon benchmarks?
\item\textbf{Q2}: Does world model guided GRPO post-training consistently improve performance?
\item\textbf{Q3}: How do subtask instruction refinement and action chunk size influence long-horizon robustness?
\item\textbf{Q4}: Can \method{} be effectively deployed in real-world robot system?
\end{itemize}

\subsection{Experimental Setup}
\noindent\textbf{Benchmark Selection.} We evaluate \method{} on two widely used simulation benchmarks, LIBERO~\cite{libero} and LIBERO-PRO~\cite{libero-pro}. LIBERO contains four main task suites: LIBERO-Spatial, LIBERO-Object, LIBERO-Goal, and LIBERO-Long. Each task suite contains 500 expert demonstrations across 10 tasks, designed to probe generalization to different spatial layouts, objects, goals, and long-horizon behaviors. LIBERO-PRO extends LIBERO with several controlled perturbation factors, enabling a more rigorous assessment of generalization and robustness under variations in objects, scene configurations, and instruction conditions. 

\noindent\textbf{Evaluation Metrics.} We report the success rate \textbf{(SR)} as the evaluation metric on both benchmarks and evaluate 50 times per task.

\noindent\textbf{Algorithmic Baselines.} We evaluate \method{} against a collection of representative and competitive baselines, including TraceVLA~\cite{tracevla}, OpenVLA~\cite{kim2024openvla}, SpatialVLA~\cite{spatialvla}, CoT-VLA~\cite{cotvla}, NORA~\cite{hung2025nora}, NORA1.5~\cite{nora15}, ThinkAct~\cite{thinkact}, $\pi_0$~\cite{pi0} and other baselines.

\noindent\textbf{Training Details.} SFT training is trained on the LIBERO dataset for 2 epochs, and evaluation runs 50 times per task. For post-training, we sample 100K trajectories and use 10 candidates for each sample. Detailed training hyperparameters are provided in the supplementary material.

\subsection{Results Analysis}


\noindent\textbf{Key Finding 1: Cross-embodiment large-scale pre-training is not necessary to achieve strong downstream performance.}
As shown in Table~\ref{tab:libero}, even without cross-embodiment large-scale pre-training, \method~ achieves an average success rate of 93\% on the LIBERO benchmark~\cite{libero}. This result indicates that training solely on in-domain multi-task demonstrations is sufficient to learn a robust vision--language-to-action mapping. In contrast, cross-embodiment large-scale pre-training not only incurs substantially higher training costs but also yields inconsistent gains on specific downstream tasks due to discrepancies in action spaces and uneven data quality.

\noindent\textbf{Key Finding 2: Post-training consistently improves performance.}
Table~\ref{tab:libero} shows that post-training improves \method{} over the SFT baseline across all LIBERO suites, with the largest gain on LIBERO-Goal (+6.0\%).
This indicates that SFT provides a strong imitation prior, but it is not sufficient to fully resolve long-horizon goal alignment. GRPO post-training addresses this gap by introducing comparative supervision at the group level over multiple candidate rollouts and using world model evaluation to reinforce trajectories that better satisfy intermediate subgoal and the final goal.
By explicitly favoring rollouts consistent with the goal while remaining anchored to demonstration-like behaviors, post-training yields more reliable goal-directed action selection, which explains the pronounced gains in goal-conditioned settings.


\noindent\textbf{Key Finding 3: \method{} exhibits strong robustness under controlled perturbations.}
Table~\ref{tab:libero_pro} shows that LIBERO-PRO~\cite{libero-pro} is substantially more challenging than standard LIBERO~\cite{libero}.
Many baselines drop to near-zero success on several perturbation dimensions, indicating limited robustness under distribution shifts.
In contrast, \method{} retains stable and non-trivial success across multiple dimensions and task suites, and continues to perform well under position and task-related perturbations.
We attribute this advantage to our two-stage training design. SFT provides a reliable imitation prior, while post-training further reinforces goal-consistent action selection under world model evaluation, improving generalization under different conditions.
\begin{table}[H]
\caption{Different Input on LIBERO-Long. \textbf{Bold} denotes the best performance among all methods.}
\centering
\small
\small
\setlength{\tabcolsep}{4pt}
\renewcommand{\arraystretch}{1.15}
\begin{tabular}{lc}
\toprule
\textbf{Input} & \textbf{LIBERO-Long} \\
\midrule
Image                                 & 80.4\% \\
Image + High-level Task               & 90.0\% \\
Image + High-level Task + Atomic Task & \textbf{92.2\%} \\
\bottomrule
\end{tabular}
\label{tab:Instruction_Mode}
\end{table}

\noindent\textbf{Key Finding 4: Fine-grained atomic subtask instruction refinement improves policy performance.}
As shown in Table~\ref{tab:Instruction_Mode}, removing language instruction causes a clear performance drop on LIBERO-Long (-9.6\%), indicating that instruction grounding is critical for robot manipulation. Conditioning the policy on the task instruction improves success to 90.0\%, and further augmenting it with a subtask description yields an additional gain to 92.2\%. We attribute this improvement to subtask instructions that provide an explicit intermediate objective for the current stage while preserving the overall task goal, helping the policy focus on the objects and actions and improving credit assignment over long-horizon tasks. 

\begin{table}[H]
\caption{Comparison of \method~ on the LIBERO benchmark with different numbers of action chunk horizon. \textbf{Bold} denotes the best performance among all methods.}
\label{tab:actionchunk}
\centering
\small
\setlength{\tabcolsep}{4pt}
\renewcommand{\arraystretch}{1.0}
\begin{tabular}{cccccc}
\toprule
Chunk Size & Spatial & Object & Goal & Long & Avg.\\
\midrule
4   & \textbf{96.4\%} & 99.6\% & 97.6\% & \textbf{94.4\%} & \textbf{97.0\%} \\
8   & 95.6\% & 99.2\% & 98.4\% & 93.0\% & 96.6\% \\
16  & 96.0\% & \textbf{99.8\%} & \textbf{98.8\%} & 91.6\% & 96.6\% \\
32  & 96.0\% & 99.0\% & \textbf{98.8\%} & 91.2\% & 96.3\% \\
\bottomrule
\end{tabular}
\end{table}

\noindent\textbf{Key Finding 5: Action chunking improves planning capability and produces more stable action sequences.}
We adopt action chunking to effectively enhance the temporal coherence and implicit planning capabilities of action generation. As shown in Table \ref{tab:actionchunk}, we empirically evaluate various chunk sizes and find that setting the size to 4 yields the optimal manipulation performance. An overly large chunk size reduces execution flexibility and exacerbates the accumulation of compounding errors, thereby degrading the overall task success rate.

\begin{table*}[t]
\caption{\normalfont Comparisons with state-of-the-art methods on the LIBERO benchmark~\cite{libero}. \textbf{Bold} denotes the best performance among all methods.}
\label{tab:libero}
\centering
\small
\setlength{\tabcolsep}{5pt}
\resizebox{\textwidth}{!}{%
\begin{tabular}{lccccccc}
\toprule
\textbf{Method} & \textbf{w/o Pre-Training} & \textbf{Model Size} & \textbf{LIBERO-Spatial} & \textbf{LIBERO-Object} & \textbf{LIBERO-Goal} & \textbf{LIBERO-Long} & \textbf{Avg.} \\
\midrule
TraceVLA~\cite{tracevla}        & \xmark & 7 B & 84.6\% & 85.2\% & 75.1\% & 54.1\% & 74.8\% \\
OpenVLA~\cite{kim2024openvla}   & \cmark & 7 B & 84.7\% & 88.4\% & 79.2\% & 53.7\% & 76.5\% \\
SpatialVLA~\cite{spatialvla}    & \cmark & 4 B & 88.2\% & 89.9\% & 78.6\% & 55.5\% & 78.1\% \\
CoT-VLA~\cite{cotvla}           & \cmark & 7 B & 87.5\% & 91.6\% & 87.6\% & 69.0\% & 83.9\% \\
$\pi_0$~\cite{pi0}              & \cmark & 4 B & 96.8\% & 98.8\% & 95.8\% & 85.2\% & 94.2\% \\
ThinkAct~\cite{thinkact}        & \xmark & 7 B & 88.3\% & 91.4\% & 87.1\% & 70.9\% & 84.4\% \\
NORA~\cite{hung2025nora}        & \cmark & 3 B & 85.6\% & 89.4\% & 80.0\% & 63.0\% & 79.5\% \\
NORA-1.5~\cite{nora15}          & \cmark & 3 B & \textbf{97.3\%} & 96.4\% & 94.5\% & 89.6\% & 94.5\% \\
\midrule
\rowcolor{red!5}
\method~ (SFT)               & \xmark & 4 B & 94.1\% & 95.4\% & 92.4\% & 90.0\% & 93.0\% \\
\rowcolor{red!5}
\method~ (SFT + Post-Training)       & \xmark & 4 B & 96.4\% & \textbf{99.6\%} & \textbf{97.6\%} & \textbf{94.4\%} & \textbf{97.0\%} \\
\rowcolor{red!8}

\rowcolor{cyan!8}

\midrule
$\Delta$ from GRPO &  &  &
\textcolor{green!60!black}{$2.3\%\uparrow$} &
\textcolor{green!70!black}{$4.2\%\uparrow$} &
\textcolor{green!60!black}{$5.2\%\uparrow$} &
\textcolor{green!60!black}{$4.4\%\uparrow$} &
\textcolor{green!60!black}{$4.0\%\uparrow$} \\
\bottomrule
\end{tabular}
}
\end{table*}

\begin{table*}[t]
\caption{\normalfont Comparisons with state-of-the-art methods on the LIBERO-PRO benchmark~\cite{libero-pro}. \textbf{Bold} denotes the best performance among all methods.}
\centering
\small
\setlength{\tabcolsep}{4.6pt}
\renewcommand{\arraystretch}{1.05}
\begin{tabular}{lcccccccccccccccccc}
\toprule
\textbf{Method}
& \multicolumn{4}{c}{LIBERO-Goal} 
& \multicolumn{4}{c}{LIBERO-Spatial} 
& \multicolumn{4}{c}{LIBERO-10} 
& \multicolumn{4}{c}{LIBERO-Object} 
& \textbf{Avg.} \\
\cmidrule(lr){2-5}\cmidrule(lr){6-9}\cmidrule(lr){10-13}\cmidrule(lr){14-17}
& Obj & Pos & Sem & Task
& Obj & Pos & Sem & Task
& Obj & Pos & Sem & Task
& Obj & Pos & Sem & Task
& \\
\midrule
$\pi_0$~\cite{pi0}       & \textbf{0.94} & 0.00 & 0.93 & 0.00 & 0.95 & 0.00 & \textbf{0.97} & 0.00 & \textbf{0.79} & 0.00 & 0.82 & 0.00 & \textbf{0.94} & 0.00 & 0.90 & 0.00 & 0.45 \\
MolmoAct~\cite{molmoact-2025} & 0.68 & 0.00 & 0.85 & 0.00 & 0.90 & 0.00 & 0.88 & 0.00 & 0.54 & 0.00 & 0.74 & 0.06 & 0.92 & 0.06 & 0.96 & 0.00 & 0.41 \\
NORA~\cite{hung2025nora}     & 0.58 & 0.00 & 0.88 & 0.00 & 0.92 & 0.00 & 0.91 & 0.00 & 0.46 & 0.00 & 0.74 & 0.00 & 0.86 & 0.00 & 0.92 & 0.00 & 0.39 \\
X-VLA~\cite{x-vla}    & 0.68 & 0.01 & \textbf{0.98} & 0.09 & \textbf{0.97} & 0.00 & 0.96 & 0.00 & 0.62 & 0.00 & 0.95 & \textbf{0.10} & 0.89 & 0.02 & 0.98 & \textbf{0.08} & 0.46 \\
\midrule
\rowcolor{red!5}
\textbf{\method~(Ours)}    & 0.81 & \textbf{0.02} & \textbf{0.98} & \textbf{0.11} & 0.95 & \textbf{0.16} & 0.95 & \textbf{0.01} & 0.55 & \textbf{0.01} & \textbf{0.95} & 0.09 & 0.93 & \textbf{0.10} & \textbf{0.99} & 0.00 & \textbf{0.48} \\
\bottomrule
\end{tabular}
\renewcommand{\arraystretch}{1.0}
\label{tab:libero_pro}
\end{table*}

\begin{figure*}[t]
    \centering
    \includegraphics[width=\linewidth]{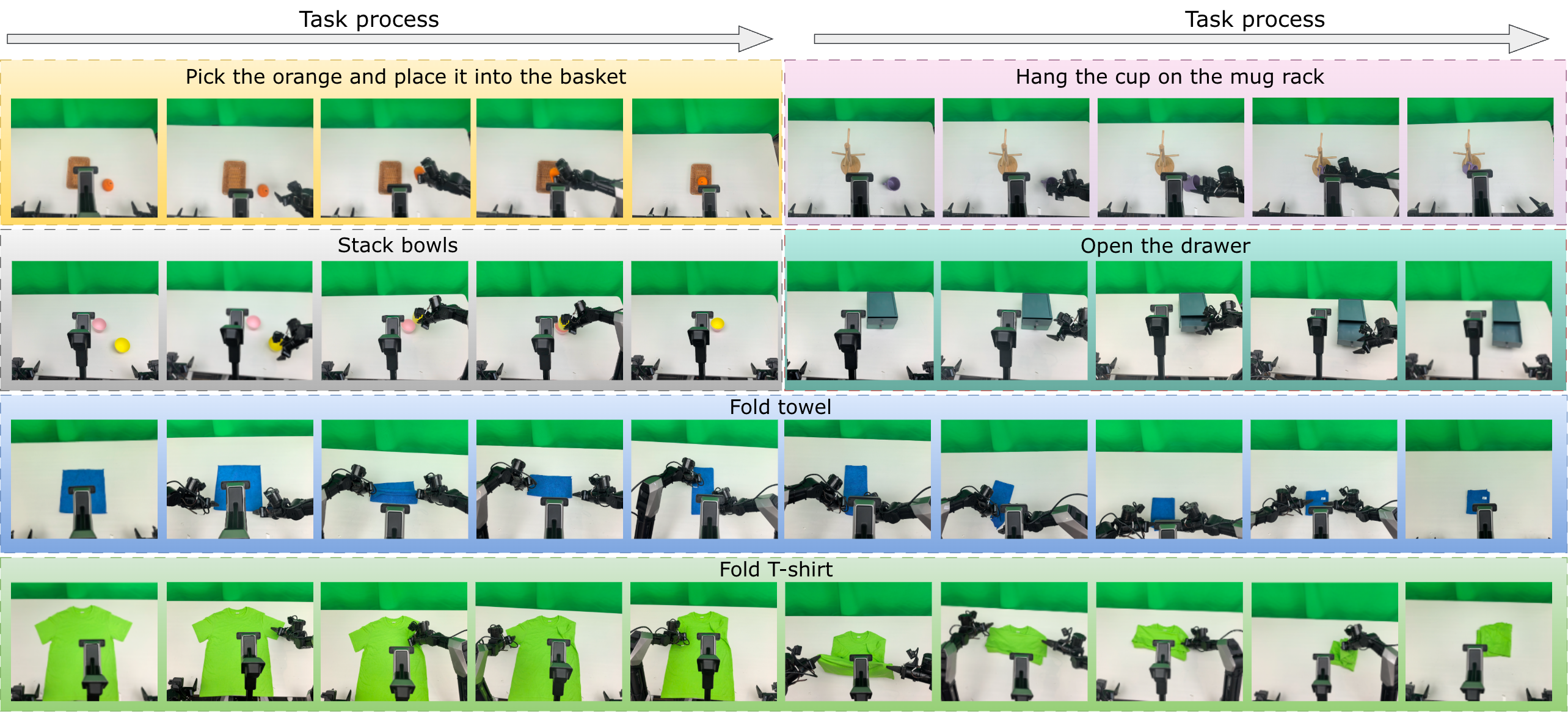}
    \caption{\textbf{Visualization of real-world tasks.} The top two rows illustrate basic tasks to stack bowls, place fruit into a basket, hang the cup, and open the drawer. The bottom two rows demonstrate hard, long-horizon tasks to fold a T-shirt and a towel.}
    \label{fig:realworld_task}
\end{figure*}
\subsection{Ablation Study}
To analyze the individual contributions of different reward elements, namely the subgoal reward $E_{\text{sub}}^{(k)}$ and the final-goal reward $E_{\text{goal}}^{(k)}$, we evaluate \method~ on the LIBERO benchmark~\cite{libero} using each elementary reward separately. The results show that both elementary rewards consistently outperform the SFT baseline, improving the average success rate by approximately 3.0\%. When adopting the full combined reward $E_{\text{sub}}^{(k)} + E_{\text{goal}}^{(k)} + D^{(k)}$, the model achieves the most stable performance, with an overall 4.0\% improvement over the SFT baseline. Notably, on the more challenging Long suite, the combined reward yields a substantial 4.4\% gain over SFT. These results suggest a clear synergy between intermediate subtask guidance and final goal consistency: the combined reward mitigates noise from purely long-horizon world model predictions while avoiding short-sighted biases induced by focusing solely on immediate subgoals.

\begin{table}[t]
\centering
\caption{\small Ablation of reward formulations on LIBERO Benchmark~\cite{libero}. \textbf{Bold} denotes the best performance among all methods.}
\label{tab:dpo_proxy_reward_ablation}
\small
\setlength{\tabcolsep}{3pt}        
\renewcommand{\arraystretch}{1.05} 
\begin{tabular}{lccccc}
\toprule
\textbf{Reward} & \textbf{Spatial} & \textbf{Object} & \textbf{Goal} & \textbf{Long} & \textbf{Avg.} \\
\midrule
SFT & 94.1\% & 95.4\% & 92.4\% & 90.0\% & 93.0\% \\
\midrule
\rowcolor{gray!12}
\multicolumn{6}{c}{\textbf{Post-Training}} \\
$E_{\text{sub}}^{(k)} + D^{(k)}$  & 96.3\% & \textbf{98.0\%} & 96.2\% & 93.5\% & 96.0\%  \\
$E_{\text{goal}}^{(k)} + D^{(k)}$ & \textbf{97.1\%} & 97.9\% & 96.0\% & 93.2\% & 96.1\%  \\
$E_{\text{sub}}^{(k)} + E_{\text{goal}}^{(k)} + D^{(k)}$ & 96.4\% & 99.6\% & \textbf{97.6\%} & \textbf{94.4\%} & \textbf{97.0\%} \\
\bottomrule
\end{tabular}
\end{table}

\begin{figure}[H]
    \centering
    \includegraphics[width=1.0\linewidth]{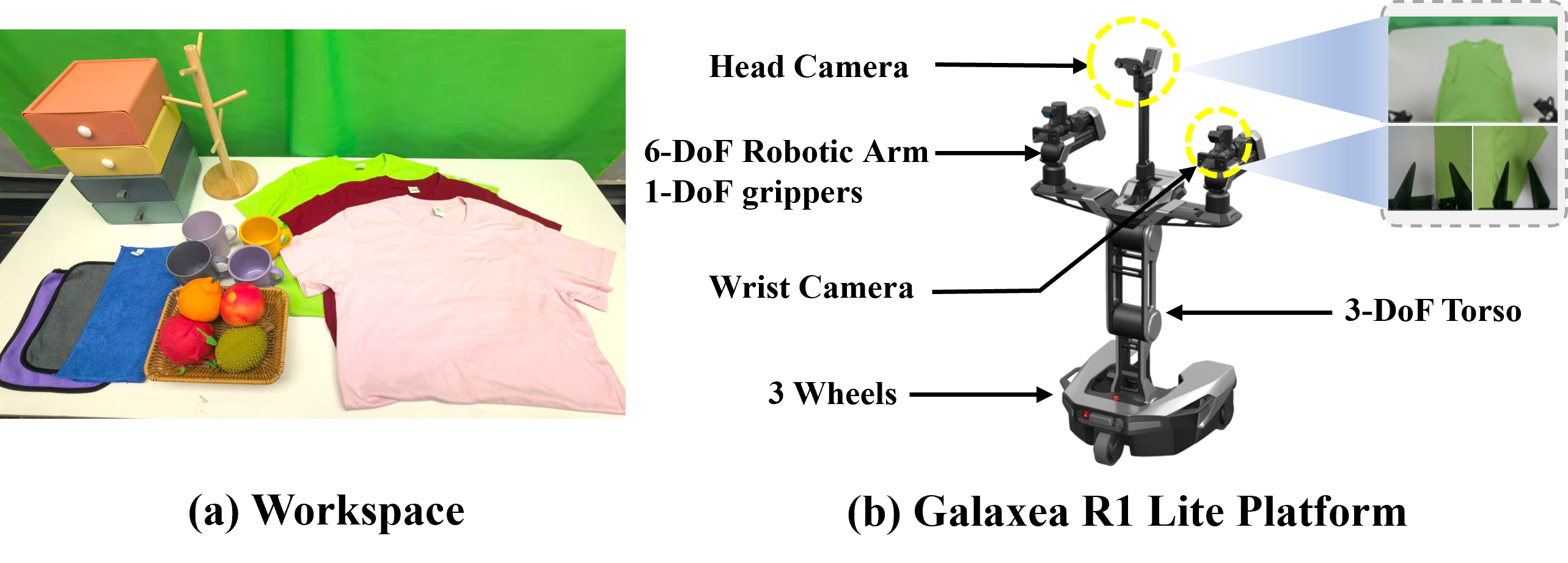}
\caption{\textbf{Real-world experimental setup.} (a) Tabletop workspace. (b) Galaxea R1 lite platform.}
    \label{fig:platform1}
\end{figure}

\section{REAL-WORLD EXPERIMENTS}
\subsection{Experimental Setup}
We set up real-world experiments based on the Galaxea R1 Lite, a dual-arm mobile platform. The system consists of three omnidirectional wheels, two 6-DoF arms, two wrist cameras, and a head camera, as shown in Figure~\ref{fig:platform1}. In our experiments, the mobile module is kept stationary. We categorize our real-world tasks into two difficulty levels: \textbf{Basic} and \textbf{Hard}. 
\textbf{Basic} tasks are short-horizon and involve basic manipulation actions; they include:

\begin{itemize}[leftmargin=*]
\item \textbf{Stack bowls}: Pick the bowl and put it on the other.
\item \textbf{Put fruit into basket}: Pick the fruit and place it into the basket.
\item \textbf{Hang cup:} Hang the cup on the mug rack.
\item \textbf{Open drawer:} Switch the drawer from closed to open.
\end{itemize}
\textbf{Hard} tasks target deformable-object manipulation and long-horizon, and they include:
\begin{itemize}[leftmargin=*]
    \item \textbf{Fold T-shirt:} Fold a T-shirt into a target folded configuration.
    \item \textbf{Fold towel:} Fold the towel twice to reach the target configuration.
\end{itemize}

\noindent\textbf{Setting.} To comprehensively assess the real-world performance of \method, we conduct evaluations under two different settings: Standard~\textbf{(ST)} and Generalization~\textbf{(GE)}. In the Standard Setting, tasks are executed under conditions consistent with the training demonstrations to verify the foundational manipulation capabilities of the learned policy. To rigorously evaluate the generalization capabilities of our framework in unstructured physical environments, the Generalization Setting introduces four distinct types of perturbations during deployment, as shown in Figure \ref{fig:platform}:
\begin{itemize}[leftmargin=*]
\item \textbf{Object Position Variation}: The initial spatial coordinates of both target objects and receptacles are randomized within the workspace.
\item \textbf{Unseen Distractor Objects}: Novel items, absent from the training demonstrations, are introduced in the vicinity of the target objects.
\item \textbf{Target Height Variation}: Receptacles are elevated to varying heights using arbitrary support structures.
\item \textbf{Instruction Variation}: Task instructions are replaced with diverse, semantically equivalent expressions.
\end{itemize}

\noindent\textbf{Training Details.} Each task comprises 100 demonstrations. The model undergoes SFT on the real-world dataset for 2 epochs, followed by post-training, where we sample 100K trajectories with 10 candidates per state. During the evaluation phase, we conduct 20 trials per task. Specifically, under the \textbf{GE} setting, the environmental conditions for these 20 trials are randomly distributed across the four types to rigorously assess robustness. Detailed hyperparameters are provided in the supplementary material.
\subsection{Results Analysis}
\noindent\textbf{Key Finding 6: \method~ demonstrates superior robustness and generalization in real-world robotic tasks.} While \method~ and the baseline $\pi_0$~\cite{pi0} exhibit comparable foundational capabilities under the \textbf{ST} setting, the critical advantage of our approach emerges under the \textbf{GE} setting. When subjected to spatial, visual, and instruction variations, the baseline performance degrades sharply to an average success rate of 29.2\%. Conversely, \method~ demonstrates substantial resilience by maintaining a 47.5\% average success rate, yielding an absolute improvement of 18.3\% over $\pi_0$~\cite{pi0}. This enhanced robustness is particularly pronounced in \textbf{Hard} tasks involving deformable objects, which require precise long-horizon control where execution errors easily compound. For instance, under the \textbf{GE} setting for the Fold T-shirt and Fold towel tasks, \method~ sustains success rates of 25\% and 35\% respectively, far exceeding the baseline. These results validate that leveraging subtask instruction decomposition and world-model-guided post-training effectively mitigates error accumulation even in highly unstructured environments.


\begin{table}[t]
\caption{Experimental results of \method~ and baseline on six real-world Galaxea R1 Lite robot manipulation tasks. \textbf{Bold} denotes the best performance among all methods.}
\label{tab:realworld1}
\centering
\setlength{\tabcolsep}{6pt}
\renewcommand{\arraystretch}{1.0}
\begin{tabular}{lcccc}
\toprule
\textbf{Task} & \multicolumn{2}{c}{$\pi_0$~\cite{pi0}} & \multicolumn{2}{c}{\textbf{\method~(Ours)}} \\
\cmidrule(lr){2-3} \cmidrule(lr){4-5}
 & \textbf{ST} & \textbf{GE} & \textbf{ST} & \textbf{GE} \\
\midrule

\rowcolor{gray!15}
\multicolumn{5}{l}{\textbf{Basic Tasks}} \\
Stack bowls           & \textbf{95\%} & 60\% & 90\% & \textbf{80\%} \\
Put fruit into basket & 85\% & 45\% & \textbf{90\%} & \textbf{65\%} \\
Hang cup              & \textbf{70\%} & 15\% & \textbf{70\%} & \textbf{35\%} \\
Open drawer           & \textbf{60\%} & 30\% & \textbf{60\%} & \textbf{45\%} \\
\midrule

\rowcolor{gray!15}
\multicolumn{5}{l}{\textbf{Hard Tasks}} \\
Fold T-shirt          & 35\% &  5\% & \textbf{40\%} & \textbf{25\%} \\
Fold towel            & 50\% & 20\% & \textbf{50\%} & \textbf{35\%} \\
\midrule

\textbf{Average}      & 65.8\% & 29.2\% & \textbf{66.7\%} & \textbf{47.5\%} \\
\bottomrule
\end{tabular}
\end{table}

\begin{figure}[t]
    \centering
    \includegraphics[width=0.99\linewidth]{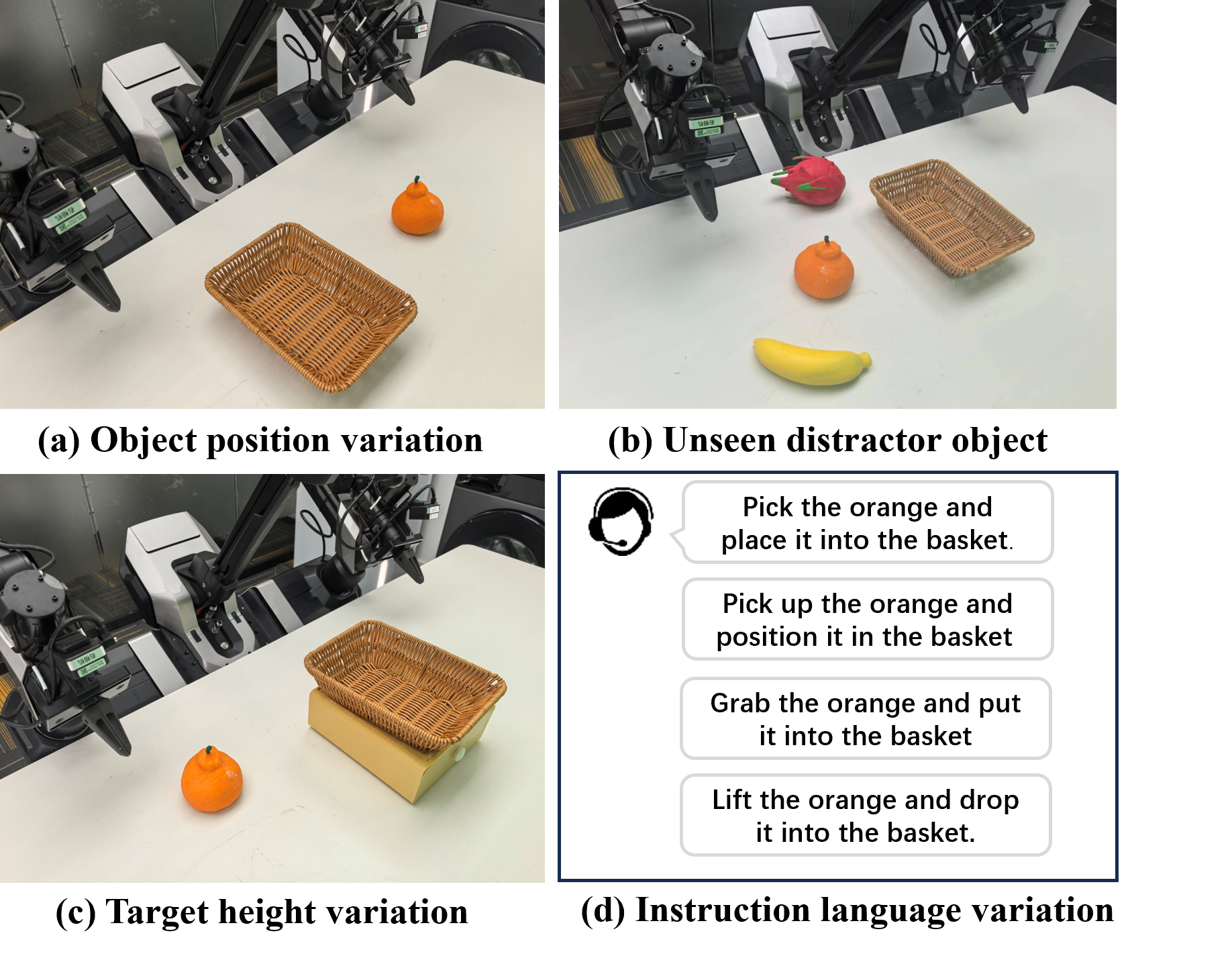}
\caption{The four types of variation introduced under the  \textbf{GE} setting to evaluate the robustness of \method. (a) Object position variation. (b) Unseen distractor object. (c) Target height variation. (d) Instruction language variation.}
    \label{fig:platform}
\end{figure}

\section{CONCLUSION}
In this paper, we propose \method~ to effectively address the instruction grounding gap and the inefficiency of offline reinforcement learning in horizon tasks. By introducing subtask decomposition driven by LLM, our training framework explicitly provides sequential guidance for complex robotic operations. Our experiments and analyses substantiate that our training paradigm, consisting of two stages, yields significant performance gains. Specifically, combining supervised fine-tuning with scalable offline GRPO guided by a predictive world model achieves remarkable results on both the LIBERO and LIBERO-PRO benchmarks. Furthermore, evaluations in the real world on the Galaxea R1 Lite platform highlight the substantial advantages of \method~ for tasks requiring prolonged execution sequences. This reliability is particularly evident in the challenging manipulation of deformable objects, such as folding a T-shirt.
While our framework enables long-horizon tasks, adapting to highly dynamic environments remains challenging because the current system relies on static subtask boundaries generated by LLM. Future work will explore end-to-end subtask generation during execution and world models that can perceive uncertainty. Furthermore, we intend to systematically enhance the generalization capabilities of our approach to handle entirely unseen objects and novel semantic instructions. 





\bibliographystyle{IEEEtran} 
\bibliography{references}
\clearpage
\onecolumn
\appendix

\section{APPENDIX}

\section{Additional Details}

This supplementary material provides additional technical details and experimental configurations to support the findings presented in the main paper. The organization of this appendix is as follows: \textbf{Section A} presents visual examples of the task segmentation process within the LIBERO benchmark, illustrating the temporal breakdown of global tasks into coarse-grained atomic sub-tasks. \textbf{Section B} details the comprehensive training and post-training hyperparameters used for both the LIBERO dataset and the real-world experiments on the Galaxea R1 Lite platform, including GPU configurations, learning rates, and input specifications. \textbf{Section C} outlines the system prompts, standardized vocabulary, and output format requirements developed to guide the model in generating precise, fine-grained atomic sub-tasks from sampled video frames.
\textbf{Section D} provides a summary of the baseline models utilized in our comparative evaluation, detailing their respective architectures, pretraining datasets, and core design principles.

\subsection{Example Figure}
\begin{figure}[h]
  \centering
  \includegraphics[width=0.99\linewidth]{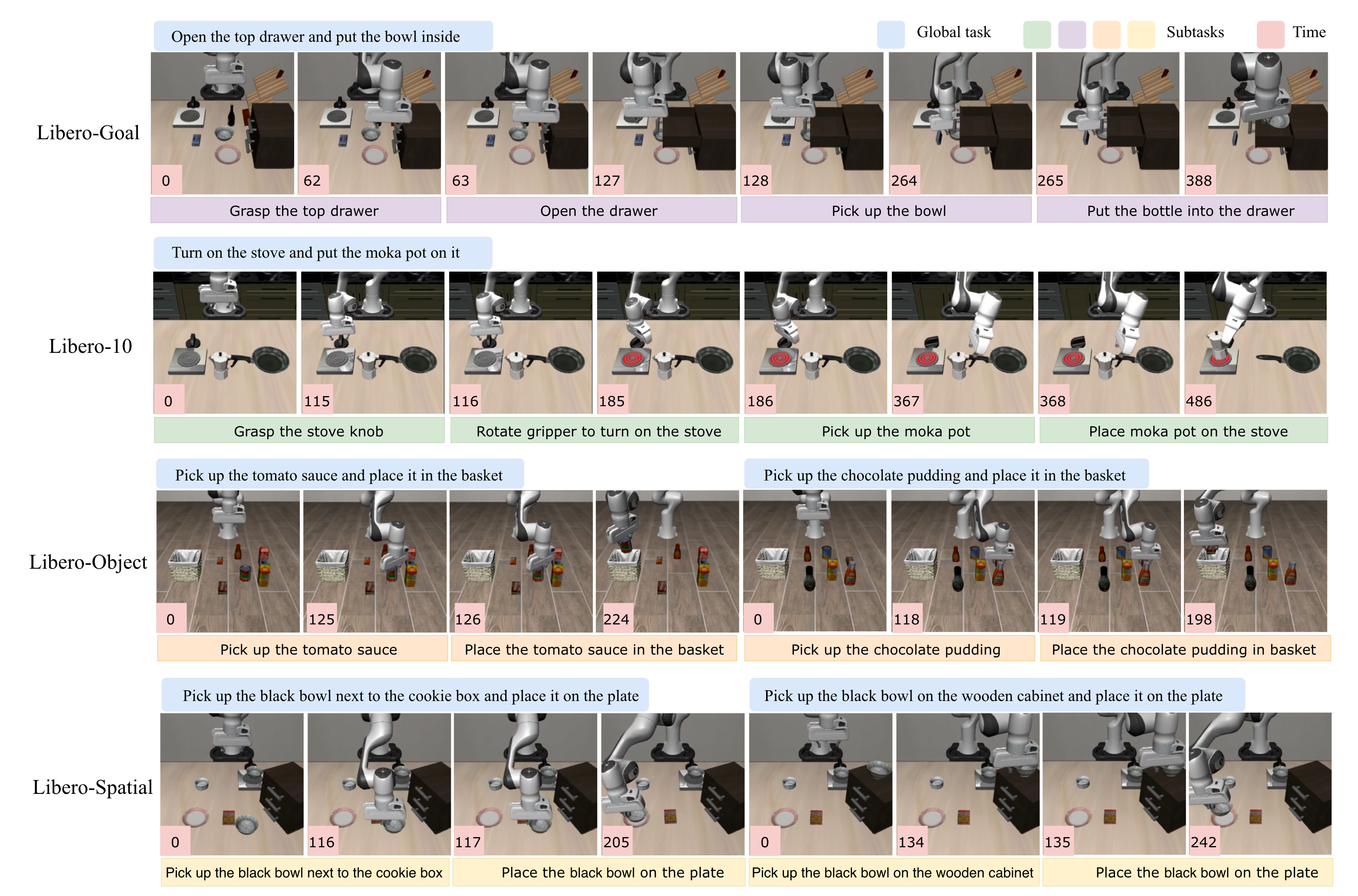}
  \caption{Atomic subtasks on the LIBERO Dataset.}
\end{figure}

\clearpage

\subsection{Training Details}
\begin{table}[H]
\centering
\caption{\normalfont SFT-Training hyperparameters for LIBERO Dataset.}
\label{tab:libero_finetune_hparams}
\small
\setlength{\tabcolsep}{6pt}
\renewcommand{\arraystretch}{1.15}
\begin{tabular}{p{0.33\textwidth} p{0.62\textwidth}}
\toprule
\textbf{Hyperparameter} & \textbf{Value} \\
\midrule
\# GPUs & $4\times$ NVIDIA H100 (80GB VRAM) \\
learning rate (LR) & $5\mathrm{e}{-4}$  \\
 batch size & 8 (per GPU) \\
\# epoch number & 2 epoch \\
input images & 1 third-person camera image, 1 wrist-mounted camera image \\
input image size & $224 \times 224$ px (wrist camera)  $224 \times 224$ px (third-person camera) \\
action chunk size ($H$) & 4 steps \\
action dimensions & 7 robot \\
\bottomrule
\end{tabular}
\end{table}

\begin{table}[H]
\centering
\caption{\normalfont Post-Training hyperparameters for LIBERO Dataset.}
\label{tab:libero_finetune_hparams}
\small
\setlength{\tabcolsep}{6pt}
\renewcommand{\arraystretch}{1.15}
\begin{tabular}{p{0.33\textwidth} p{0.62\textwidth}}
\toprule
\textbf{Hyperparameter} & \textbf{Value} \\
\midrule
\# GPUs & $2\times$ NVIDIA H100 (80GB VRAM) \\
 batch size & 8 (per GPU) \\
input images & Only 1 third-person camera image \\
input image size & $224 \times 224$ px \\
sample number & 100k \\
Candidate & 10 \\
action dimensions & 7 robot \\
\bottomrule
\end{tabular}
\end{table}

\begin{table}[H]
\centering
\caption{\normalfont SFT-Training hyperparameters for Real-World Dataset on Galaxea R1 Lite platform.}
\label{tab:libero_finetune_hparams}
\small
\setlength{\tabcolsep}{6pt}
\renewcommand{\arraystretch}{1.15}
\begin{tabular}{p{0.33\textwidth} p{0.62\textwidth}}
\toprule
\textbf{Hyperparameter} & \textbf{Value} \\
\midrule
\# GPUs & $4\times$ NVIDIA H100 (80GB VRAM) \\
learning rate (LR) & $5\mathrm{e}{-4}$  \\
 batch size & 8 (per GPU) \\
input images & 1 head camera image, 2 wrist camera image \\
input image size & $1280 \times 720$ px (head camera)  $640 \times 360$ px (wrist camera) \\
action chunk size ($H$) & 10 steps \\
action dimensions & 14 robot \\
\bottomrule
\end{tabular}
\end{table}

\begin{table}[H]
\centering
\caption{\normalfont Post-Training hyperparameters for Real-World Dataset on Galaxea R1 Lite platform.}
\label{tab:libero_finetune_hparams}
\small
\setlength{\tabcolsep}{6pt}
\renewcommand{\arraystretch}{1.15}
\begin{tabular}{p{0.33\textwidth} p{0.62\textwidth}}
\toprule
\textbf{Hyperparameter} & \textbf{Value} \\
\midrule
\# GPUs & $2\times$ NVIDIA H100 (80GB VRAM) \\
 batch size & 8 (per GPU) \\
input images & Only 1 head camera image \\
input image size & $1280 \times 720$ px \\
sample number & 100k \\
Candidate & 10 \\
action dimensions & 14 robot \\
\bottomrule
\end{tabular}
\end{table}

\subsection{Prompt for fine-grained atomic task generation}
\begin{tcolorbox}[
    colback=gray!10!white,
    colframe=gray!50!black,
    title=Prompt for Instruction Generation,
    fonttitle=\bfseries,
    boxrule=0.5mm,
    arc=2mm,
    width=\columnwidth,
    breakable,
    before skip=2mm,
    after skip=2mm,
    left=3pt,
    right=3pt,
    top=3pt,
    bottom=3pt
]
{\ttfamily\small
You are an expert in robotic manipulation analysis for tabletop tasks (Libero benchmark). 

I will provide frames from a robot manipulation video.

\vspace{2mm}
Your mission is to act as an expert Vision-Language Model providing precise temporal segmentations for tabletop robotic manipulation. You will generate a contiguous sequence of coarse-grained atomic sub-tasks (aligned with the ``GLOBAL TASK CONTEXT'') based on SAMPLED VIDEO FRAMES, FRAME INDICES, and the OVERALL GOAL.

You need to identify and sequence all the main steps that constitute the episode using the STANDARDIZED VOCABULARY defined below, ensuring complete temporal coverage with no overlapping or missing frames.

\vspace{2mm}


    


\vspace{2mm}

\noindent\textbf{<Global Task Context>}\\
The overall goal of the robot in this video is: \texttt{"task\_instruction"}
    
 Please use this context to correctly interpret the robot's actions. Segment the video into COARSE-GRAINED atomic sub-tasks that achieve this goal.

\vspace{2mm}

\noindent\textbf{<Granularity Rules (Coarse-Grained)>}
\begin{itemize}[label=-]
    \item  Merge "Approach" + "Grasp" -> "Pick up".\\
    \item Merge "Move" + "Release" -> "Place".\\
    \item A typical pick-and-place episode consists of 2-4 main steps.
\end{itemize}

\vspace{2mm}

\noindent\textbf{<Standardized Vocabulary>}
\begin{itemize}[label=-]
    \item "Pick up [object]"\\
    \item "Place [object] into/on [target]"\\
    \item "Move arm to [location]"\\
    \item "Open/Close [object]"\\
    \item "Push [object]"\\
\end{itemize}    
Be specific with object names (e.g., "white mug", "left plate") matching the Global Task Context.

\vspace{2mm}

\noindent\textbf{<Output Format>}\\
Strictly return a JSON object:
\begin{verbatim}
    {
        "tasks": [
            {
                "instruction": "Pick up the red cola can", 
                "start_frame": 0, 
                "end_frame": 45
            },
            {
                "instruction": "Place the red cola can into the white basket", 
                "start_frame": 46, 
                "end_frame": 120
            }
        ]
    }
\end{verbatim}
}
\begin{itemize}[label=-]
\item Note: The provided images are sampled. Image 1 corresponds to the frame index provided in the user prompt. Use these indices to estimate start/end frames accurately.
\end{itemize}  

\end{tcolorbox}

\clearpage

\subsection{Baseline.}
The following baselines are utilized for comparative evaluation.

\noindent\textbf{OpenVLA:} A 7B-parameter VLA model that integrates a Llama 2 language backbone with a hybrid visual encoder fusing pretrained representations from DINOv2 and SigLIP. The model is pretrained on a large-scale collection of 970k real-world robot trajectories sourced from the Open X-Embodiment dataset.

\noindent\textbf{TraceVLA:} A spatial-temporal enhanced VLA model that incorporates visual trace prompting to encode state–action trajectories into the visual input. Built upon OpenVLA and fine-tuned on 150K robot manipulation trajectories.

\noindent\textbf{Octo-base:} A transformer-based generalist manipulation policy trained on approximately 800K trajectories from the Open X-Embodiment dataset. It supports language and goal-image conditioning and is designed to accommodate diverse sensor modalities and action spaces across multiple robotic platforms.

\noindent\textbf{SpatialVLA:} A spatially-aware VLA model designed to enhance 3D understanding in robot manipulation. Pretrained on 1.1M real-world robot episodes, it introduces Ego3D Position Encoding to inject explicit 3D spatial information into visual observations and employs Adaptive Action Grids to discretize and represent spatial robot movements for cross-robot transfer.

\noindent\textbf{CoT-VLA:} A 7B VLA model that incorporates explicit visual chain-of-thought reasoning for manipulation. It autoregressively predicts future visual frames as intermediate goals and then generates short action sequences to reach them, enabling temporal planning. 

\noindent\textbf{$\pi_0$:} A VLA model that builds a flow-matching action architecture on top of a pretrained VLM to inherit large-scale semantic knowledge. It is trained on diverse datasets collected from multiple dexterous robotic platforms. The model supports direct language prompting and fine-tuning for new skills, demonstrating broad task coverage and cross-platform generalization.

\noindent\textbf{ThinkAct:} A dual-system VLA framework that separates high-level reasoning from low-level control. It trains a multimodal LLM to generate embodied reasoning plans, which are reinforced using action-aligned visual rewards and compressed into a visual latent plan. This latent representation conditions a downstream action policy, enabling improved long-horizon planning, few-shot adaptation, and robust execution in complex embodied manipulation tasks.

\noindent\textbf{NORA:} A 3B-parameter VLA model designed for efficient real-time robotic manipulation. Built on the Qwen-2.5-VL-3B backbone, it leverages strong visual-semantic understanding to improve action grounding while significantly reducing computational overhead compared to larger VLA models. It trained on 970K real-world robot demonstrations and equipped with the FAST+ tokenizer for efficient action sequence generation.

\noindent\textbf{NORA-1.5:} An enhanced VLA model built upon the pre-trained NORA backbone, incorporating a flow-matching–based action expert to improve reliability and task performance. It uses reward-driven post-training with action-conditioned world models and deviation-from–ground-truth heuristics, enabling direct preference optimization for adaptation to new embodiments.

\noindent\textbf{MolmoAct:} An Action Reasoning Model that integrates depth-aware perception, mid-level spatial planning, and low-level action prediction. The 7B-D variant achieves strong zero-shot, fine-tuned, and out-of-distribution performance across simulation and real-world tasks, surpassing prior VLAs.

\noindent\textbf{X-VLA:} A flow-matching–based VLA model that uses soft-prompt embeddings to capture cross-embodiment variations in heterogeneous robotic datasets. Its 0.9B variant, X-VLA-0.9B, achieves state-of-the-art performance across simulations and real-world robots, enabling flexible dexterity, fast adaptation, and effective exploitation of diverse robotic platforms with minimal additional parameters.

\end{document}